%% file: main.tex
\documentclass[letterpaper, 10 pt, conference]{ieeeconf}  

\input{preamble}

\title{\LARGE \bf Geometry-Aware Predictive Safety Filters on Humanoids: \\ From Poisson Safety Functions to CBF Constrained MPC}

\author{Ryan M. Bena, Gilbert Bahati, Blake Werner, Ryan K. Cosner, Lizhi Yang and Aaron D. Ames%
\thanks{This work was supported by BP and the Technology Innovation Institute.}%
\thanks{The authors are with the Department of Mechanical and Civil Engineering, California Institute of Technology, Pasadena, CA 91125 USA. Email:  {\tt\small \{ryanbena, gbahati, bwerner, rkcosner, lzyang, ames\} @caltech.edu}.}%
}

\begin{document}

\maketitle
\thispagestyle{empty}
\pagestyle{empty}

\begin{abstract}
Autonomous navigation through unstructured and dynamically-changing environments is a complex task that continues to present many challenges for modern roboticists. In particular, legged robots typically possess manipulable asymmetric geometries which must be considered during safety-critical trajectory planning. This work proposes a \textit{predictive safety filter}: a nonlinear \textit{model predictive control} (MPC) algorithm for online trajectory generation with geometry-aware safety constraints based on \textit{control barrier functions} (CBFs). Critically, our method leverages \textit{Poisson safety functions} to numerically synthesize CBF constraints directly from perception data. We extend the theoretical framework for Poisson safety functions to incorporate temporal changes in the domain by reformulating the static Dirichlet problem for Poisson's equation as a parameterized moving boundary value problem. 
Furthermore, we employ Minkowski set operations to lift the domain
into a configuration space that accounts for robot geometry. 
Finally, we implement our real-time predictive safety filter on humanoid and quadruped robots in various safety-critical scenarios. The results highlight the versatility of Poisson safety functions, as well as the benefit of CBF constrained model predictive safety-critical controllers.
\end{abstract}

\section{INTRODUCTION}

Robotic technology is evolving at an exciting rate; with this rapid evolution comes the increasing need for safe robotic autonomy. Whether one considers a quadrupedal robot performing an industrial inspection, an autonomous vehicle transporting passengers, or the Ingenuity helicopter flying above the Martian surface, collisions with the environment can often be irreversible and catastrophic. Humanoid robots in particular are designed to operate in complex and dynamically-changing human-occupied environments. Thus, it is critical to ensure that autonomous robots interact with their environments in a safe, predictable, and robust manner.

The field of safety-critical control aims to tackle this challenge. Over the past several decades, various approaches have been proposed to address the notion of safe autonomy for dynamical systems. A few popular examples include Hamilton-Jacobi (HJ) reachability methods \cite{bansal_hj_2017}, artificial potential fields (APFs) \cite{singletary2021comparative}, state-constrained model predictive control (MPC) \cite{borrelli_mpcBook_2017}, and control barrier function (CBF) methods \cite{ames2019control}.

Of particular interest in this work are MPC and CBFs. MPC has demonstrated repeated success in robotics, with results ranging from legged locomotion \cite{grandia2023perceptive} to autonomous car racing \cite{rosolia2019learning}. Meanwhile, CBFs continue to gain traction as an intuitive and efficient means of enforcing dynamic safety constraints on a broad class of systems \cite{zhang2020haptic,breeden2022space,molnar2025collision}. Furthermore, the merger of MPC and CBFs, denoted MPC+CBF, has recently been highlighted as a powerful safety-critical control architecture \cite{zeng_dtcbfMpc_2021,grandia2021multi,roque2022corridor}. MPC+CBF methods leverage the finite-horizon optimality of the traditional MPC problem with the dynamic safety guarantees of CBF constraints. Specifically, when CBF constraints are used along the MPC horizon, the resultant closed-loop autonomous system has provable robustness and recursive feasibility benefits \cite{cosner2025dynamic}.

\begin{figure}
\label{fig: hero figure}
    \centering    
    \includegraphics[width=\linewidth]{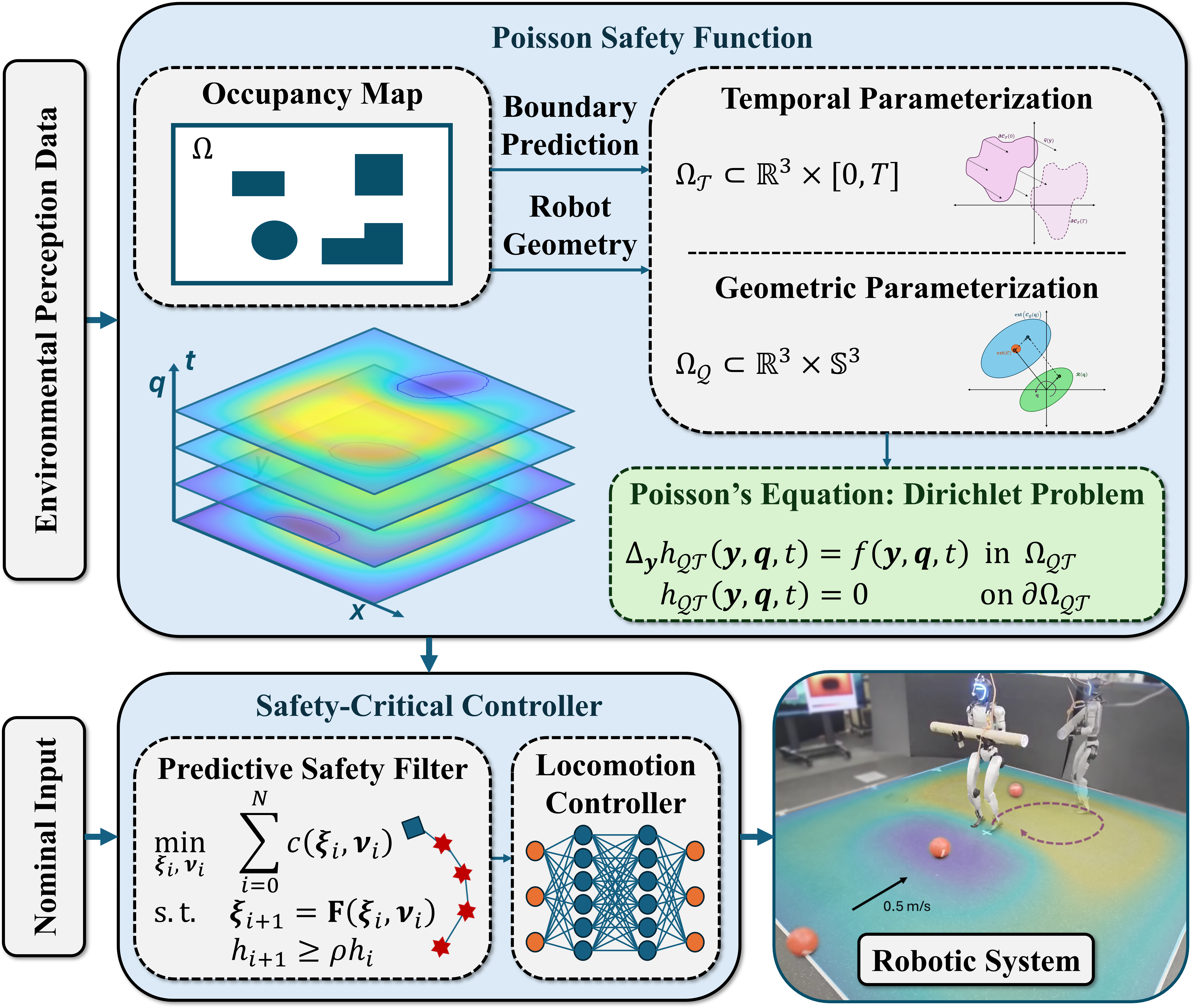}
    \vspace{-2ex}
    \caption{Predictive Safety Filtering with Poisson Safety Functions. From environmental perception data, we formulate a Dirichlet problem for Poisson's equation with temporal and configurational dependencies. The resultant \textit{Poisson safety function} informs the safety constraints within our CBF+MPC \textit{predictive safety filter}. This safety filter generates a reference trajectory for an RL locomotion controller, which produces joint-level torque commands for a legged robot, resulting in safe autonomous behavior. Experimental footage is available at \url{https://youtu.be/i8uMyW4iSQw}.}
    \vspace{-3ex}
\end{figure} 

MPC+CBF methods seek to perform some nominal autonomous task while keeping the state of the robotic system within a safe region. This is often formalized as enforcing forward invariance of a \textit{safe set} with respect to the closed-loop dynamics. As the first step in achieving forward invariance, the safe set is frequently characterized by a continuous \textit{safety function}, the zero superlevel set of which precisely defines the safe set. In most published work, the safety function is specified via two methods: 1) an \textit{ad hoc} analytical expression --- this is the case in most CBF research, where ``obstacles" in the environment are described by implicit geometric shapes, e.g., circles, ellipses, and rectangles\cite{glotfelter2020nonsmooth,molnar2023composing}, or 2) a signed distance function (SDF) --- this approach characterizes safety via a numerical function that represents the distance to the nearest safe set boundary (i.e. obstacle surface) \cite{oleynikova2017voxblox,long2021learning}. We explore a more recent result: the \textit{Poisson safety function} \cite{bahati2025poisson}, a numerical method which uses partial differential equations (PDEs) to characterize safe sets by solving Poisson's equation in the spatial domain. Poisson safety functions have demonstrated benefits over \textit{ad hoc} and SDF methods. 

In this work, we present a geometry-aware \emph{predictive safety filter} for robotic systems, based on nonlinear MPC+CBF, that utilizes Poisson safety functions in higher-dimensional configuration space to enforce safe set forward invariance via translational and rotational degrees-of-freedom (DOFs). The main contributions of this paper are as follows:

\begin{itemize}
    \item Recasting the static Dirichlet problem in \cite{bahati2025poisson} as a moving boundary problem, we generate a time-varying safety function that extrapolates the evolution of the safe set over a finite future time interval. 
    \item Using Minkowski set operations based on robot occupancy, we lift the domain for Poisson's equation into higher-dimensional configuration space, thus producing a Poisson safety function that has additional DOFs.
    \item Employing our extended Poisson safety function as a CBF for a single-integrator, we formulate a novel MPC+CBF predictive safety filter, which enables safe geometry-aware trajectory generation in real-time.
    \item We experimentally demonstrate the advantages of our online predictive safety filtering method on quadruped and humanoid robots in multiple challenging safety-critical scenarios.
\end{itemize}

\section{PRELIMINARIES}
\label{sec: Preliminaries}

\subsection{Safety-Critical Control: Control Barrier Functions}

In this work, we discuss safety for control affine dynamical systems, i.e., systems whose state $\bs{x}\in\R^n$ evolves according to the continuous-time dynamics: 
\begin{equation}
\label{eq: Affine Dynamics OL}
    \dot{\bs{x}} = \bs{f}(\bs{x}) + \bs{g}(\bs{x})\bs{u},
\end{equation}

\noindent where $\bs{f}:\R^n\rightarrow\R^n$ and $\bs{g}:\R^n\rightarrow\R^{n\times m}$, assumed to be Lipschitz continuous, characterize the autonomous and non-autonomous behavior of the system, respectively, and $\bs{u}\in\R^m$ is the control input. Given some Lipschitz continuous feedback law $\bs{u}=\bs{k}(\bs{x})$, the closed-loop dynamics are:
\begin{equation}
\label{eq: Affine Dynamics CL}
    \dot{\bs{x}} = \bs{f}(\bs{x}) + \bs{g}(\bs{x})\bs{k}(\bs{x}).
\end{equation}

%
%
%
%
    
%
We characterize the safety of the system \eqref{eq: Affine Dynamics OL} with respect to a set $\Sc$:
\begin{align}
\label{eq: Safe Set S}
    \Sc \coloneqq \left\{\bs{x}\in\R^n \,\big|\, h(\bs{x}) \geq 0 \right\},\\
    \partial\Sc \coloneqq \left\{\bs{x}\in\R^n \,\big|\, h(\bs{x}) = 0 \right\},\\
    \mathrm{int}\left(\Sc\right) \coloneqq \left\{\bs{x}\in\R^n \,\big|\, h(\bs{x}) > 0 \right\},
\end{align}

\noindent where the zero superlevel set of the function $h: \R^n \rightarrow \R$ represents the subset of $\R^n$ for which the system state is considered safe. 
The aim of safety-critical control is to design a feedback control law $\bs{k}(\bs{x})$ that renders $\Sc$ forward invariant.

\begin{definition}

    (Forward Invariance)  A set $\Sc$ is \textit{forward invariant} with respect to \eqref{eq: Affine Dynamics CL} if, for every initial state $\bs{x}(t_0) \in \Sc$, the resulting state trajectory $t \mapsto \bs{x}(t)$ remains in $\Sc$ for all $t\geq t_0$.
    
\end{definition}

For control affine systems as in \eqref{eq: Affine Dynamics OL}, the forward invariance of $\Sc$ can be enforced using CBFs.

\begin{definition}
    (Control Barrier Function (CBF) \cite{ames2017control})
    Let $\Sc \subset \R^n$ be a set as in \eqref{eq: Safe Set S}, with function $h$ satisfying $\nabla h(\bs{x}) \neq 0$ when $h(\bs{x}) = 0$. Then, the function $h$ is a CBF on $\Sc$ if there exists $\gamma \in \mathcal{K}_{\infty}^e$ such that for all $\bs{x} \in \R^n$:
    \begin{align}
    \label{eq: CBF}
    \!\!\!\! \sup_{\bs{u} \in \R^m} \left\{\dot{h}(\bs{x},\bs{u}) = L_{\bs{f}} h(\bs{x}) + {L_{\bs{g}} h(\bs{x})} \bs{u} > -\gamma(h(\bs{x})) \right\}.
    \end{align}
\end{definition}

Given a CBF $h$ and function $\gamma$, the set of control inputs satisfying \eqref{eq: CBF} is given by:
\begin{align}
    \mathcal{U}_\mathrm{CBF}(\bs{x}) = \left \{\bs{u} \in \R^m \, \big | \,  \dot{h}(\bs{x}, \bs{u})  \geq -\gamma(h(\bs{x})) \right\},
\end{align}

\noindent such that any locally Lipschitz controller for which $\bs{u}\in\mathcal{U}_\mathrm{CBF}(\bs{x})$ results in the forward invariance of $\Sc$ \cite{ames2017control}.


For real-world robotic systems with zero-order hold, sampled-data implementations it is generally impossible to choose continuous control actions that create the closed-loop system \eqref{eq: Affine Dynamics CL}. Instead, for the remainder of this work we focus on the discrete time formulations of these systems:
\begin{align} \label{eq: Discrete Dynamics}
    \bs{x}_{k+1} & = \mb{F}(\bs{x}_{k}, \bs{u}_k),  & \forall k \in \mathbb{N}, 
\end{align}
where $\mb{F}: \R^{n} \times \R^m  \to \R^{m}$ is the discretization of \eqref{eq: Affine Dynamics OL} over a finite time interval $\delta_t >0$ for a constant input, $\bs{u}$. 
Given this version of the system, we focus on the forward invariance of the discrete-time system at sample times\footnote{\vspace{-1em}See \cite{breeden_sampledDataCbfs_2022} for a discussion of zero-order-hold, intersample safety.} $k \delta_t$ which can be achieved using the discrete-time formulation of the CBF: 
\begin{definition}[Discrete-time Control Barrier Functions (DCBF) \cite{agrawal_dtcbf_2017}]
    Let $\Sc$ be the set given in \eqref{eq: Safe Set S} and $\rho \in [0,1]$. A function $h: \R^n \to \R$ is a discrete-time control barrier function (DCBF) for \eqref{eq: Discrete Dynamics} if 
    for each $\bs{x}\in \Sc$ 
    there exists a $\bs{u} \in \R^m $ such that: 
    \begin{align}
        h(\mb{F}(\bs{x}, \bs{u})) \geq \rho h(\bs{x}). 
    \end{align}
\end{definition}

This discrete-time formulation can be rewritten as: 
\begin{align}
\label{eq: DCBF 2}
    h(\mb{F}(\bs{x}, \bs{u})) - h(\bs{x}) \geq (\rho - 1) h(\bs{x}),
\end{align}
where the change in safety, represented by the left-hand side of \eqref{eq: DCBF 2}, is constrained by a function of the previous value, as in the case of the continuous time CBF constraint \eqref{eq: CBF}.
Similar to the safety guarantees of CBFs, DCBFs have a related safety guarantee for the discrete-time states \cite[Prop. 4]{agrawal_dtcbf_2017} where the system is kept safe at sampling times $k$ since the value of safety is lower bounded by a geometrically decaying curve, i.e. $h(\bs{x}_k) \geq \rho^k h(\bs{x}_0)$. 

DCBFs can be used to generate safe control actions in the form of a safety filter, where performance goals are encoded via a nominal, but potentially unsafe, controller $\bs{\pi}_\textup{nom}: \R^n \to \R^m $ which is minimally modified in a point-wise fashion to produce safe control inputs: 
\begin{align}
    \bs{\pi}_\textup{cbf}(\bs{x}) = \underset{\bs{u}\in \R^m }{\textup{arg min}} & \quad \Vert \bs{u} - \bs{\pi}_\textup{nom}(\bs{x})\Vert^2\\
    \textup{s.t.} & \quad h(\mb{F}(\bs{x}, \bs{u})) \geq \rho h(\bs{x}).
\end{align} 

In practice, the \textit{safe set} that we wish to render forward invariant is often specified via desired outputs $\bs{y}: \re^n \rightarrow \re^p$, 
%
\begin{align}
\label{eq: Safe Set}
    \Cc \coloneqq \left\{\bs{y}\in\R^p \,\big|\, h_0(\bs{y}) \geq 0 \right\},
\end{align}

\noindent where the zero superlevel set of the \textit{safety function} $h_0: \R^p \rightarrow \R$ represents the subset of the output space for which the system is considered safe. 

\begin{remark}
The safe set $\C$ and the aforementioned set $\Sc$ are not necessarily the same set; similarly, the safety function $h_0$ and the CBF $h$ are not necessarily the same function. For certain outputs $\bs{y}$, $h$ may be equal to $h_0$, but generally $h$ must be derived from $h_0$ via some dynamic extension method (i.e., Exponential CBFs \cite{nguyen_ecbf_2016}, High Order CBFs \cite{xiao2022high}, CBF Backstepping \cite{taylor2022backstepping},
etc.). The need for dynamic extension often depends on the \textit{relative degree} \cite{isidori1985nonlinear} of the outputs $\bs{y}$ with respect to the inputs $\bs{u}$. 
\end{remark}

\subsection{Model Predictive Control}


Despite the theoretical utility of CBFs in guaranteeing safety, they can often result in undesirable closed-loop properties in their practical implementation when the performance goals, encoded in the nominal controller, conflict with the safety constraints. This can result in the emergence of undesirable stable equilibrium points which keeps the system safe, but prevents it from achieving its goal \cite{mestres2025control}. 
To simultaneously achieve both safety and performance goals, it is often useful to incorporate horizon-based optimization into the controller. To do this we adopt an MPC framework that incorporates the DCBF constraint as in \cite{zeng_dtcbfMpc_2021,cosner2025dynamic}, enforcing safety over a time horizon. 

The \emph{predictive safety filter} is defined using the following finite-time optimal control problem (FTOCP) with a DCBF contraint along a horizon of length $N \in \mathbb{N}_{\geq 1 }$:
\begin{align}
    \underset{\bs{\nu}_{0:N-1} \in \R^m }{\underset{\bs{\xi}_{0:N} \in \R^n }{\min }} & \quad \sum_{i=0}^{N-1} c(\bs{\xi}_i, \bs{\nu}_i) + V(\bs{\xi}_N) \\
    \textup{s.t. } & \quad \bs{\xi}_{i+1} = \mb{F}(\bs{\xi}_i, \bs{\nu}_i), \quad \forall i \in \{ 0, \dots, N-1\}  \nonumber\\
    & \quad h(\bs{\xi}_{i+1}) \geq \rho h(\bs{\xi}_i), \quad \forall i \in \{ 0, \dots, N-1\} \nonumber\\
    & \quad \bs{\xi}_0 = \bs{x}_k \nonumber 
\end{align}

\noindent where $c: \R^n \times \R^m \to \R$ is the stage cost, $V: \R^n \to \R$ is the terminal cost used to approximate the infinite-horizon optimal control problem, and $\bs{\xi}_i \in \R^n $ and $\bs{\nu}_i \in \R^m $ represent the planned state and inputs at time $k+i$. The first constraint ensures that the planned trajectory is dynamically feasible, the second constraint is the DCBF condition that enforces safety along the trajectory, and the last constraint ensures that the plan aligns with the current state. 

To generate an input using this FTOCP, the optimal plan of inputs is computed as $[\bs{\nu}_0^*(\bs{x}_k), \dots, \bs{\nu}_{N-1}^*(\bs{x}_k) ]$, and $\bs{\nu}_0^*(\bs{x}_k)$ is applied to the system, defining the MPC+CBF controller: 
\begin{align}
    \bs{\pi}^{\textup{MPC+CBF}}(\bs{x}_k) = \bs{\nu}_0^*(\bs{x}_k).  
\end{align}
By enforcing the safety constraint in the FTOCP and optimizing across the planning horizon $N$, the $\bs{\pi}^\textup{MPC+CBF}$ controller displays favorable performance and robustness properties when compared to standard MPC or CBF controllers \cite{cosner2025dynamic}. 

\section{EXTENDING POISSON SAFETY FUNCTIONS}
\label{sec: Poisson}

\subsection{Spatial Poisson Safety Functions}

In robotic collision avoidance, the safety function $h_0$ used to characterize $\C$, as in \eqref{eq: Safe Set}, is traditionally defined as a function of the spatial output $\bs{y}\in\R^3$. However, the nature of the physical spatial environment is frequently unknown \textit{a priori}, and thus it can be difficult to predetermine an analytical expression for $h_0$. More often, general occupancy information for the local environment is collected in real-time, either via onboard sensors or global perception systems. This occupancy information can be used to estimate membership of the safe set $\Cc$, without an explicit functional characterization. In such cases, the estimated boundary of $\Cc$, i.e. $\partial\Cc$, can be used to define the zero level set of the function $h_0$. Subsequently, this level set can be used to determine the entire function over some bounded domain via a Dirichlet problem for Poisson's equation \cite{bahati2025poisson}. 

Let $\C$ define a domain: 
\begin{align}
    \Occ \coloneqq \Cc, \quad \partial\Omega \coloneqq \partial\C, \; \text{and} \quad \Oc \coloneqq \mathrm{int}(\C),
\end{align}

\noindent where $\Omega\subset\R^3$ is the smooth, open, bounded, and connected set over which Poisson's equation will be solved; $\partial\Omega$ is the boundary of that set; and $\Occ$ represents the closure of $\Oc$, i.e. $\Occ=\Omega \cup \partial\Omega$. On this domain, we can establish the following Dirichlet problem for Poisson's equation:

\begin{equation}
\label{eq: Poisson}
    \left\{
    \begin{aligned}
        \Delta h_0(\bs{y}) &= f(\bs{y}) \quad&\forall\bs{y}\in\Omega, \\
        h_0(\bs{y}) &= 0 \quad &\forall\bs{y}\in\partial\Omega,
    \end{aligned}
    \right.
\end{equation}

\noindent where $\Delta=\nabla\cdot\nabla$ is the \textit{Laplace} operator, which computes the divergence of the gradient of $h_0$; the forcing term $f: \Omega \rightarrow \R_{<0}$ is a user-defined function that forces $h_0$ to be superharmonic. In \cite{bahati2025poisson}, it was shown that, if the forcing function $f$ is smooth on the domain, i.e., $f\in C^{\infty}(\Occ)$, then the resultant Poisson safety function is also smooth on that domain, i.e., $h_0\in C^{\infty}(\Occ)$. It follows that $h_0$ will be a CBF for systems with single-integrator dynamics; furthermore, $h_0$ is compatible with the aforementioned CBF dynamic extension methods.

\subsection{Temporal Parameterization --- Boundary Prediction}
\label{sec: T Extension}

\begin{figure*}
    \centering    \includegraphics[width=\linewidth]{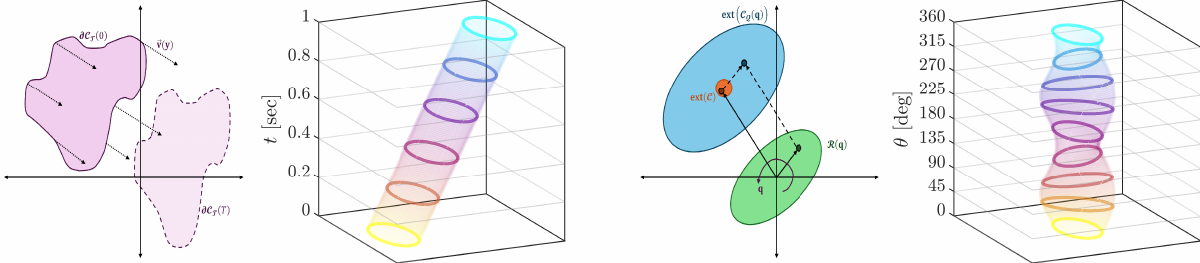}
    \vspace{-1ex}
    \caption{Extended Non-Cylindrical Domains. \textbf{[Left]} Temporal forward prediction. The cartoon shows how the vector field $\bvv$ can be used to linearly extrapolate the future set boundary $\partial\C_\T$. \textbf{[Middle Left]} Example of a temporally extended non-cylindrical domain for a 2D robot, i.e. $\Oct\subset\R^2\times[0,T]$. \textbf{[Middle Right]} The Minkowski sum/difference operation. The diagram shows an illustration of the Minkowski difference in \eqref{eq: Minkowski}.  \textbf{[Right]} Example of a geometrically extended non-cylindrical domain for a 2D robot with a single rotational DOF, i.e. $\Ocq\subset\R^2\times\Sp^1$. The robot geometry is represented by an ellipse. As the value of $\theta$ changes, the resultant Minkowski difference produces a non-cyclindrical domain on which Poisson's equation must be solved.}
    \label{fig: geometric poisson}
    \vspace{-3ex}
\end{figure*} 

In previous research, discussions of safe sets and Poisson safety functions leveraged the assumption that safe sets do not change explicitly with respect to time. In dynamic environments however, the set itself can change over time, and thus its corresponding safety function will have an explicit time dependence. To account for these temporally-evolving environments, we can modify the set $\C$, leading to $\C_\T(t)\subset\R^3$, where $t\in[0,T]$ is the value of time, defined over some bounded interval. 

Determining the time dependence of $\C_\T$ (and the corresponding membership of the temporally lifted set $\Omega_\T\subset\R^3\times[0,T]$) requires knowledge of the behavior of the safe set $\C_\T(t)$ and its boundary $\partial\C_\T(t)$ for $t \geq 0$. In practice, robotic systems often observe their environments in real-time; thus, the evolution of $\C_\T$ may not be known \textit{a priori}. In such cases, it becomes necessary to predict the time evolution of $\C_\T$.
Level set methods \cite{osher2004level} are widely utilized techniques for propagating boundaries by representing them implicitly and modeling their evolution using appropriate partial differential equations. The boundary $\partial\C_\T(t)$ is represented as the zero level set of an implicit function as follows:
\begin{equation}
\partial\C_\T(t) = \{\by \in \re^3 : \phi(\by, t) = 0  \}.
\end{equation}
\noindent
Since $ \phi(\by, t) = 0$ on the evolving front, by the chain rule, we arrive at the transport equation:

\begin{equation}\label{eq: level set equation}
\frac{\partial \phi}{\partial t}(\by, t) + \bvv(t) \cdot\nabla \phi (\by, t)=0,
\end{equation}

\noindent with the initial condition $\phi(\by, 0) = \phi_0(\by)$, where $\dot{\by}(t) = \bvv(t)$ for each point $\by(t) \in \partial\C_\T(t)$. The values of the solution to \eqref{eq: level set equation} remain constant along characteristic curves \cite{evans2022partial}, defined as solutions to the ODE $\dot{\by}(t) = \bvv(t)$ with $ \by(0) \in \partial\C_\T(0)$. This Lagrangian formulation allows us to track the boundary's evolution over time \cite{hieber2005lagrangian}, enabling us to determine $\partial\C_\T(t)$ for each $t \in [0,T]$ by rewriting $\phi(\by,t)$ as:
\begin{equation}
    \phi(\by(t),t) = \phi_0 \left(\by(t) - \int_0^t\bvv(\tau)d\tau\right),
\end{equation}
\noindent where $\bvv$ is assumed to be Lipschitz continuous in its arguments. This leads to the prediction model:
\begin{equation}
\label{eq: prediction NL}
\partial\C_\T(t) = \left\{\by \in \re^3 : \phi_0 \left(\by - \int_0^t\bvv(\tau)d\tau\right)=0  \right\},
\end{equation}
 where we assume the variation of $\partial\C_\T(\cdot)$ is continuous in time, following from the appropriate regularity of $\phi$.  

Finally, we can construct the following non-cylindrical space-time domain: 
\begin{align}
    \Occ_\T = \bigcup_{t \in [0,T]} \Cc_\T(t) \times \{t\} \subset\R^3\times[0,T],
\end{align}
%
%
which represents a temporally-lifted set corresponding to $\C_\T(t)$, allowing us to formulate the moving boundary Dirichlet problem for Poisson's equation as follows:
\begin{equation}
\label{eq: T Poisson}
    \left\{
    \begin{aligned}
        \Delta_{\bs{y}} h_\T(\bs{y},t) &= f(\bs{y},t) \quad &\forall(\bs{y},t)\in\Omega_\T, \\
        h_\T(\bs{y},t) &= 0 \quad &\forall (\bs{y},t)\in\partial\Omega_\T,
    \end{aligned}
    \right.    
\end{equation}

\noindent where $\Delta_{\bs{y}}=\frac{\partial}{\partial\bs{y}} \cdot \frac{\partial}{\partial\bs{y}}$ represents the \textit{Laplacian} operator computed strictly with respect to the spatial output $\bs{y}$, and the Poisson safety function $h_\T : \R^3\times[0,T] \rightarrow \R$ characterizes the safe set $\C_\T$.
Given a constant velocity field $\bvv$, Fig.\;\ref{fig: geometric poisson} shows an example of the non-cylindrical domain $\Oc_\T$ representing a moving obstacle as time evolves. 

In equation \eqref{eq: T Poisson}, we regard $h_\T$ as a collection of functions of space $\bs{y}$ parametrized by $t$. The regularity of solutions to \eqref{eq: T Poisson} have been addressed in $\cite{bonaccorsi2001variational,bogelein2021evolutionary}$ in the general case for parabolic equations, under the condition that the boundary evolves continuously in time, as we assumed.

In what follows, we generalize this parametric dependence of the domain to other settings relevant for robotic applications. In particular, paralleling the method by which we extend the safe set $\C$ to account for time-varying environments, we modify $\C$ to consider the physical geometry of the robotic system, and its relationship to the robot's rotational DOFs. 

\subsection{Geometric Parameterization --- Safe Set Buffering}
\label{sec: Q Extension}

Given the safe set $\C$ in \eqref{eq: Safe Set}, a point is considered safe as long as its spatial output $\bs{y}\in\R^3$ is in $\C$. Consequently, physical robotic platforms are often controlled as a point mass via their centroidal motion, and thus the techniques described in Section\;\ref{sec: Preliminaries} are applied with respect to some arbitrary point on the robot. However, real-world robots have asymmetric geometries which must be suitably considered during collision avoidance. A common naive approach to handling robot geometry is to define a single robot \textit{radius} and reduce the safe set $\C$ by this radius equally in all spatial dimensions. Although this method is often computationally efficient, it lacks awareness of the true shape of the robot, which can result in significant conservatism, especially when dealing with highly asymmetric robot geometries. 

Instead, we introduce the concept of \textit{safe set buffering}, where the safe set (representing free space in the environment) is modified to account for the potentially-complex geometric shape of the robotic system in a particular orientation. First, we define the set $\mathcal{R}(\bs{q})\subset\R^3$ of all points occupied by the robot, relative to its centroid, where $\bs{q}\in\Sp^3$ is a quaternion representing the attitude of the system with respect to some global reference frame. In this work, we model the robot as a rigid-body, but in general the robot geometry might be further manipulable via additional DOFs in some high-order configuration space.

Next, we construct the reduced safe set $\C_{\mathcal{Q}}(\bs{q})\subset\C$. This is accomplished via a \textit{Minkowski difference}\footnote{This is equivalent to the \textit{Minkowski sum}: $\mathrm{ext}\left(\C_\Q(\bs{q})\right) = \mathrm{ext}\left(\C\right) \oplus \mathcal{R}(\bs{q})$. In practice, the Minkowski difference can be computed as a convolution of the occupancy map with a kernel representing the robot occupancy.}:
\begin{align}
\label{eq: Minkowski}
    \C_\Q(\bs{q}) = \C \ominus \mathcal{R}(\bs{q}).
\end{align}

\noindent The resultant safe set $\C_\Q(\bs{q})$ is the minimal pointwise reduction of $\C$ required to keep the entire robotic platform collision-free. An illustration of the Minkowski difference operation is presented in Fig.\;\ref{fig: geometric poisson}.

Re-examining the PDE formulation in \eqref{eq: Poisson}, the new attitude-dependent safe-set $\C_\Q(\bs{q})$ can be lifted to a higher dimensional space to create the new non-cylindrical domain:
\begin{align}
    \Occ_\Q = \bigcup_{\bq \in \mathbb{S}^3} \C_\Q(\bq) \times \{\bq\} \subset\R^3\times\Sp^3,
\end{align}
which is extended to incorporate robot orientation. We can solve for the rotationally-dependent Poisson safety function:
\begin{equation}
\label{eq: Q Poisson}
\left\{
    \begin{aligned}
        \Delta_{\bs{y}} h_\Q(\bs{y},\bs{q}) &= f(\bs{y},\bs{q}) \quad &\forall(\bs{y},\bs{q})\in\Ocq, \\
        h_\Q(\bs{y},\bs{q}) &= 0 \quad &\forall(\bs{y},\bs{q})\in\partial\Ocq.
    \end{aligned}\right.
\end{equation}
\noindent Because the original safe set $\C$ only defines safety with respect to the translational output $\bs{y}$, the rotational output vector $\bs{q}$ does not need to satisfy a boundary condition along $\bs{q}$ and thus does not need to appear in the Laplacian. 
The regularity properties for solutions to \eqref{eq: Q Poisson} follow similar arguments as in the previous subsection.
% ...

As an example, a non-cylindrical domain $\Ocq$ for Poisson's equation is depicted in Fig.\;\ref{fig: geometric poisson}. In the figure, we reduce the dimensionality of $\Ocq\subset\R^2\times\Sp^1$, allowing us to visually examine how the set $\C_\Q$ changes when evaluated at different values of $\theta$, and how this impacts the lifted set $\Ocq$.

\section{CONTROL ARCHITECTURE}
\label{sec: Controller}

Using the methods in Sections\;\ref{sec: T Extension} and \ref{sec: Q Extension}, let $\Ocqt$ be a non-cylindrical domain in configuration space and time:
\begin{align}
    \!\!\!\Occ_{\Q\T} \!= \!\bigcup_{\bq,\,t} \C_{\Q\T}(\bq,t) \times \{\bq\} \times \{t\} \subset\R^3\times\Sp^3\times[0,T],
\end{align}
\noindent where $\C_{\Q\T}(\bs{q},t)$ is a safe set parameterized in orientation and time. This allows us to formulate the final parameterized Dirichlet problem for Poisson's equation:
\begin{equation}
\label{eq: QT Poisson}
\left\{
    \begin{aligned}
        \Delta_{\bs{y}} h_{\Q\T}(\bs{y},\bs{q},t) &= f(\bs{y},\bs{q},t) \quad \!\!\!\!&\forall(\bs{y},\bs{q},t)\in\Ocqt, \\
        h_{\Q\T}(\bs{y},\bs{q},t) &= 0 \quad \!\!\!\!&\forall(\bs{y},\bs{q},t)\in\partial\Ocqt.
    \end{aligned}
\right.
\end{equation}

\begin{figure}
    \centering    \includegraphics[width=\linewidth]{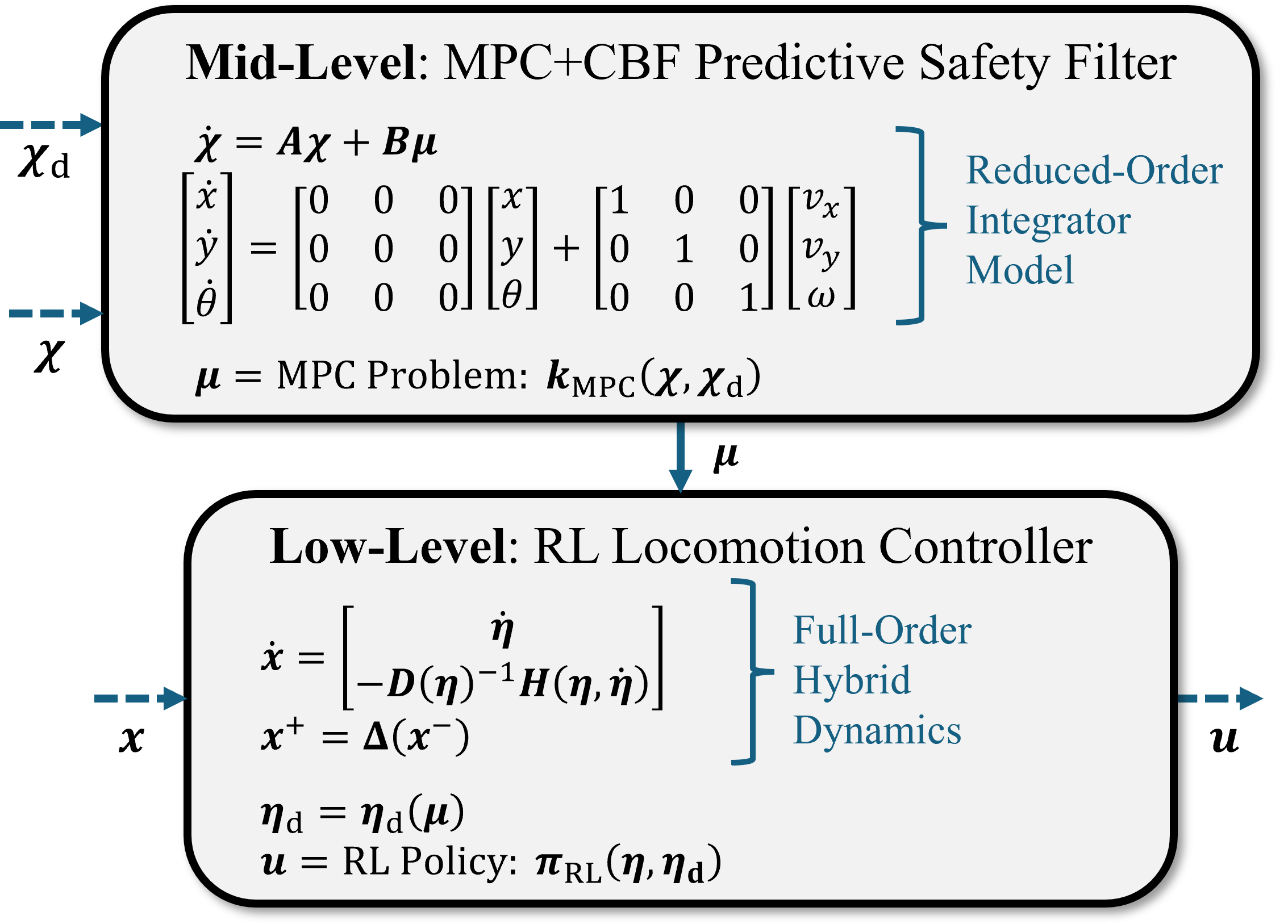}
    \vspace{-1ex}
    \caption{Layered Control Architecture. Our mid-level MPC+CBF predictive safety filter provides the velocity references for the low-level RL locomotion controller. Safety can be guaranteed for the full-order dynamics, as long as the low-level controller produces ``sufficient" tracking performance \cite{cohen2024rom}.}
    \label{fig: control architecture}
    \vspace{-3ex}
\end{figure}

\noindent 
Solving this PDE produces the Poisson safety function $h_{\Q\T}$ corresponding to the safe set $\C_{\Q\T}(\bs{q},t)$. In \cite{bahati2025poisson}, it was proven that the static Poisson safety function $h_0$ in \eqref{eq: Poisson} is a CBF for robotic systems with single-integrator dynamics.  While safety cannot be formally guaranteed in general time-varying environments, due to the inherently unknown nature of the future temporal behavior of $\C_{\Q\T}$, we can still harness $h_{\Q\T}$ to perform safety-critical control using an ROM-based hierarchical framework \cite{tamas2022model, cohen2024rom}, as depicted in Fig.\;\ref{fig: control architecture}.

We employ a ROM for predictive safety filtering; specifically, we represent the system as a fully-actuated single-integrator with two translational DOFs and a single rotation DOF. The resultant 3-DOF continuous-time system model is $\dot{\bs{\chi}}=\bs{\mu}$, where the reduced-order state $\bs{\chi}=[x,y,\theta]^\top\in\R^2\times\Sp^1$ is directly controlled via the input $\bs{\mu}=[v_x,v_y,\omega]^\top\in\R^3$.

With this model, we design an MPC-based feedback control policy based on discrete-time single-integrator dynamics, which nominally drives the system state $\bs{\chi}$ to the goal state $\bs{\chi}_\mathrm{d}$ while enforcing the forward invariance of $\C_{\Q\T}$ using DCBF constraints:
\begin{align}
\label{eq: NMPC}
    \underset{\bs{\xi}_i,\;\bs{\nu}_i}{\min } & \quad \sum_{i=0}^N \left(\bs{\chi}_\mathrm{d}-\bs{\xi}_i\right)^\top\bs{Q}(\bs{\chi}_\mathrm{d}-\bs{\xi}_i) + \bs{\nu}_i^\top\bs{R}\bs{\nu}_i \\
    \textup{s.t. } & \quad \bs{\xi}_{i+1} = \bs{\xi}_i + \bs{\nu}_i\delta_t, \quad \forall i\in[0,N-1]  \nonumber\\
    & \quad h_{\Q\T}(\bs{\xi}_{i+1},t_{i+1}) \geq \rho h_{\Q\T}(\bs{\xi}_i,t_i), \quad \forall i \in [0,N-1] \nonumber\\
    & \quad \bs{\xi}_0 = \bs{\chi}_k. \nonumber 
\end{align}
\noindent In our MPC formulation, the quadratic costs $\bs{Q}$ and $\bs{R}$ are chosen to elicit the desired system behavior; the time step $\delta_t\in\R_{>0}$ is chosen in conjunction with $N$ to yield a suitable prediction horizon $T$; the CBF parameter $\rho\in\left(0,1\right)$ is selected to produce appropriate dynamic robustness; and the initial $\bs{\xi}_0$ is constrained to match the current state $\bs{\chi}_k$.

The MPC problem in \eqref{eq: NMPC} is a nonlinear nonconvex optimization problem, which we solve using sequential quadratic programming (SQP). The first control output in the SQP minimizer $\bs{\nu}_0^*$ is applied as the current control action $\bs{\mu}_k$. While $\bs{\nu}_0^*$ is not necessarily optimal with respect to the original problem \eqref{eq: NMPC}, in practice it is sufficient to achieve the desired system-level performance.

\begin{figure*}
    \centering    \includegraphics[width=\linewidth]{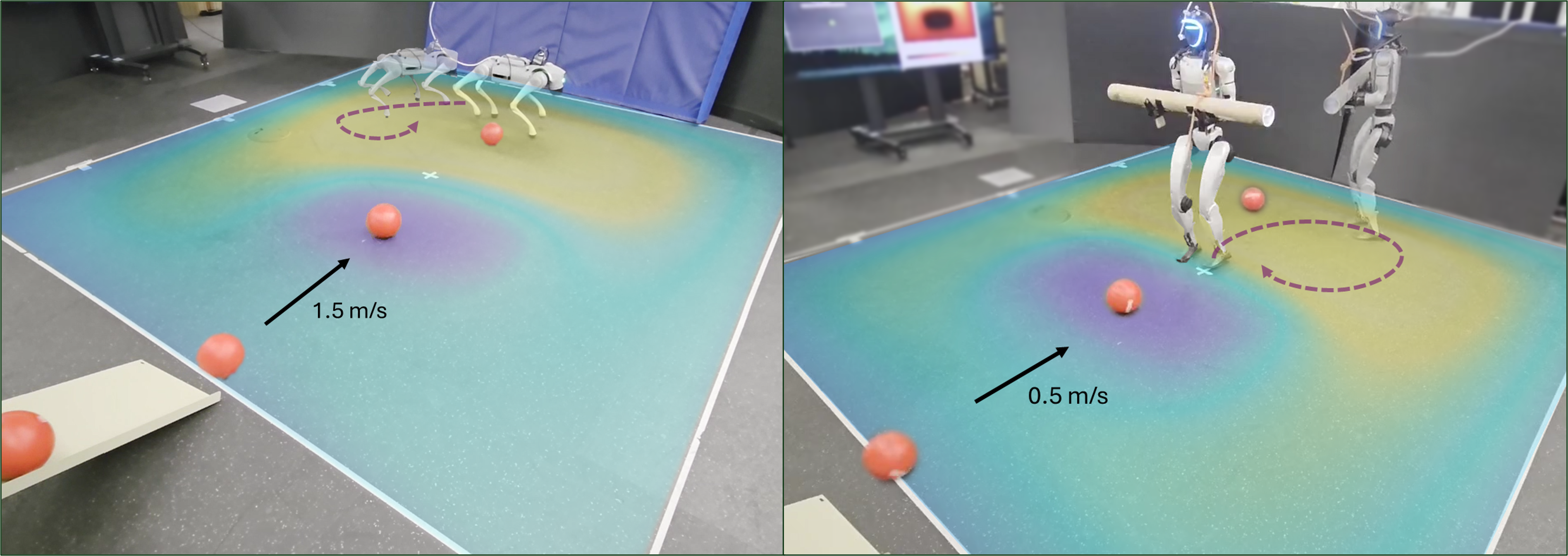}
    \vspace{-2ex}
    \caption{Dynamic Collision Avoidance Scenario. Composite image showing a single avoidance maneuver for each robotic platform: Go2 quadruped (left) and G1 humanoid (right). By considering the relationship between orientation and robot geometry during the safety filtering process, the legged robots avoid the dynamic obstacles by simultaneously translating and rotating. Footage of additional trials is available in the supporting video at \url{https://youtu.be/i8uMyW4iSQw}.}
    \label{fig: quadruped experiment photo}
\end{figure*}

\begin{figure*}
    \centering    \includegraphics[width=\linewidth]{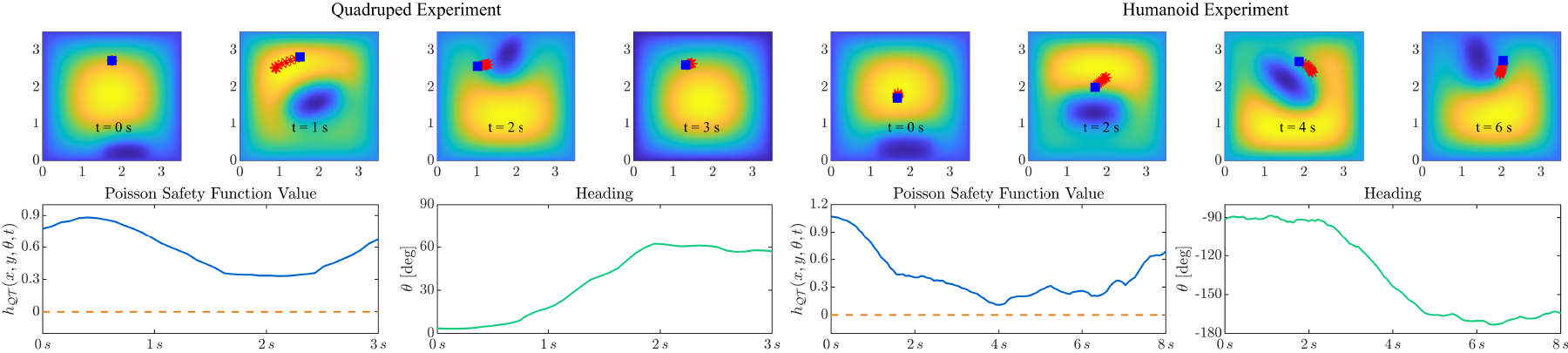}
    \vspace{-2ex}
    \caption{Dynamic Collision Avoidance Data. \textbf{[Top]} Manifolds of the Poisson safety function $h_{\Q\T}$. The depicted color maps represent manifolds of $h_{\Q\T}$, evaluated at specific times and corresponding orientations, throughout each quadrupedal and humanoid avoidance maneuver. The resultant MPC trajectories are overlaid in red. \textbf{[Bottom Far-Left]} The evaluated value of $h_{\Q\T}$ throughout the quadruped avoidance. This value is positive, verifying that safety was maintained. \textbf{[Bottom Mid-Left]} The measured heading angle of the quadruped. The quadrupedal robot pivots away from the incoming dynamic obstacle. \textbf{[Bottom Mid-Right]} The evaluated value of $h_{\Q\T}$ throughout the humanoid avoidance. This value is positive, verifying that safety was effectively maintained. \textbf{[Bottom Far-Right]} The measured heading angle of the humanoid. As with the quadruped, the humanoid robot pivots away from the dynamic obstacle.}
    \label{fig: quadruped experiment data}
    \vspace{-3ex}
\end{figure*} 

When integrated with our legged robotic platforms, the velocity command $\bs{\mu}$ is relayed to a low-level locomotion controller. For the humanoid robot, this low-level controller consists of a walking policy trained in IsaacLab \cite{mittal2023orbit} using standard rewards from \cite{gu2024advancing}. The training is conducted with domain-randomized forces on the robot's upper body to emulate forces from upper body twists and carried loads. A separate optimization-based controller is applied to the waist in order to effectively track rotational velocity commands without compromising the robot's balance. We implement the following QP controller, which optimally tracks the nominal heading rate command $\omega$ from \eqref{eq: NMPC} through a combination of lower-body rotation rate $\omega_\alpha\in\R$ and relative upper-body rotation rate $\omega_\beta\in\R$, while simultaneously reducing any upper-body misalignment $\beta_0\in\Sp^1$:
\begin{align}
    \omega_\alpha^*, \omega_\beta^* = \underset{\omega_\alpha, \omega_\beta}{\textup{argmin}} &  \quad \Tilde\omega^2 + \lambda\beta^2 \\
    \textup{s.t.} & \quad \Tilde\omega = \omega_\alpha + \omega_\beta - \omega \nonumber \\ & \quad \beta = \beta_0 + \omega_\beta \delta_t \nonumber \\
    & \quad \omega_{\alpha,\text{min}} \leq \omega_\alpha \leq \omega_{\alpha,\text{max}}. \nonumber
\end{align}

\noindent The lower-body angular velocity command $\omega_\alpha$ is then sent to the RL controller, while the upper-body command $\omega_\beta$ is tracked with a PD controller. 

\section{IMPLEMENTATION}

\begin{figure*}
    \centering    \includegraphics[width=\linewidth]{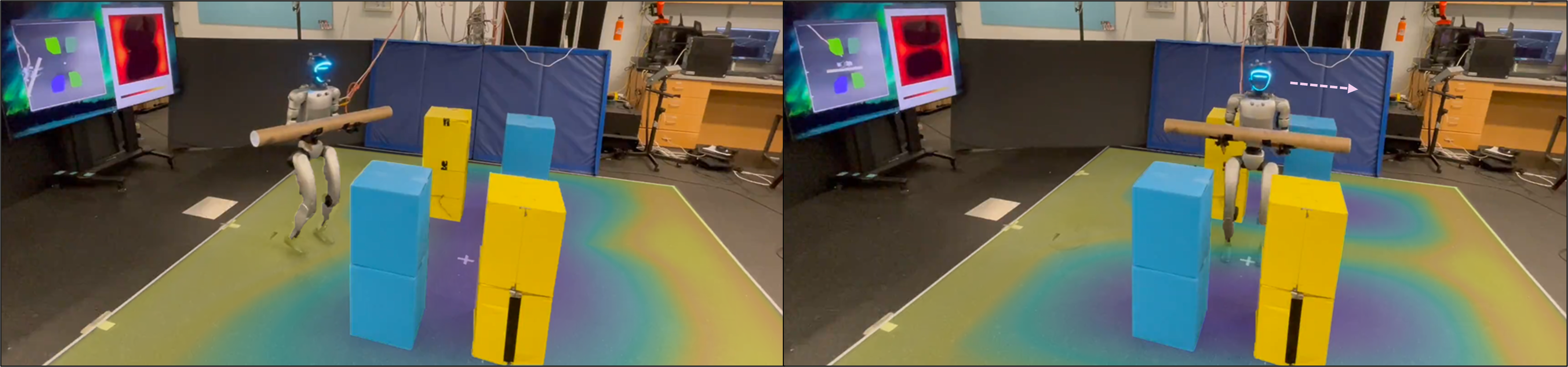}
    \vspace{-2ex}
    \caption{Environmental Navigation Scenario. Composite image showing the humanoid traversing a narrow corridor while carrying an oblong payload at $t=50\,s$ (left) and $t=60\,s$ (right). Footage of the entire experiment is presented in the supporting video.}
    \label{fig: humanoid experiment photo}
    \vspace{-1em}
\end{figure*} 

\begin{figure}
    \centering    \includegraphics[width=\linewidth]{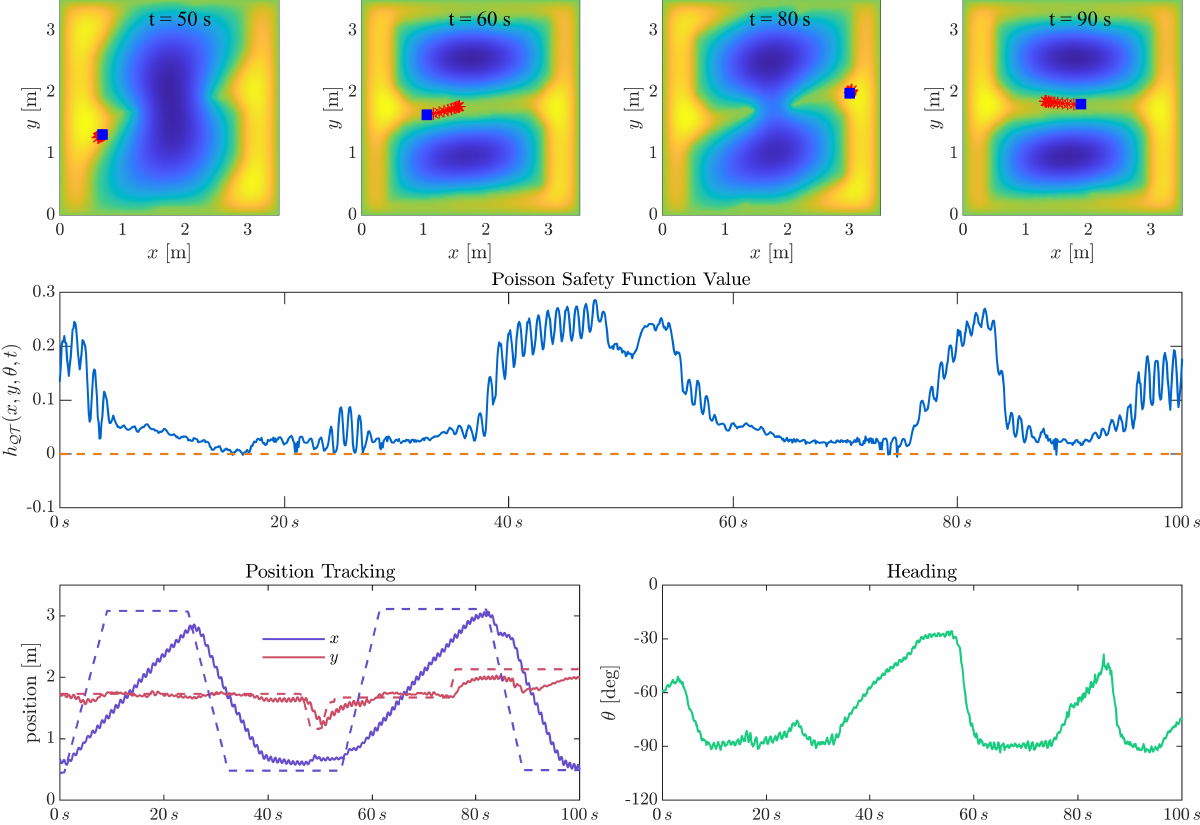}
    \vspace{-2ex}
    \caption{Environmental Navigation Data. \textbf{[Top]} Manifolds of the Poisson safety function $h_{\Q\T}$, as computed at four distinct times during the 100-second navigation task, evaluated at the current time and orientation. The corresponding MPC trajectory is overlaid in red. \textbf{[Middle]} The evaluated value of $h_{\Q\T}$ throughout the experiment. This value is positive, verifying that safety was maintained. \textbf{[Bottom Left]} The measured position and corresponding reference (dashed). The humanoid robot passed back and forth between the narrow corridor to reach its goal. \textbf{[Bottom Right]} The measured heading angle of the humanoid. The robot maneuvers its heading to maintain safety while passing between the obstacles.}
    \label{fig: humanoid experiment data}
    \vspace{-3ex}
\end{figure} 

\subsection{Experimental Setup}

We implemented our proposed safety-critical controller on two legged robotic platforms (the Unitree Go2 quadruped and G1 humanoid) in multiple collision avoidance scenarios. The humanoid was made to carry an oblong payload to simulate a potential industrial application for which rotational considerations are paramount. For both experimental implementations, we estimated robot pose using an OptiTrack motion capture system, and we capture environmental perception data with an overhead ZED 2i stereo camera. The overhead camera stream was passed to an efficient track-anything-model (eTAM) image segmentation model \cite{xiong2024efficient}, which identified and tracked user-selected objects in the experimental space. The segmented image was directly used as a 2D occupancy map, indicating free vs. occupied space. 

To model the evolution of an occupancy map directly from perception data, we employed an OpenCV \textit{optical flow} algorithm \cite{farneback2003two}. Given an RGB camera image, the optical flow method estimates the velocity of each pixel, resulting in a 2D velocity field $\bvv$. Taking the assumption of constant $\bvv$ over the MPC prediction horizon, we used \eqref{eq: prediction NL} to derive the linear boundary prediction model\footnote{More complex environmental prediction models could further enhance performance. This is an area of active research.}:
\begin{equation}
\label{eq: prediction L}
\partial\C_\T(t) = \left\{\by \in \re^2 : \phi_0 \left(\by - \bvv t\right)=0  \right\}.
\end{equation}
Next, we performed the Minkowski difference in \eqref{eq: Minkowski} for a discrete sampling of robot geometries along $\Sp^1$. We generated a domain for Poisson's equation $\Omega_{\Q\T}$ and solved the parameterized Dirichlet problem in \eqref{eq: QT Poisson} using a successive overrelaxation (SOR) method \cite{strikwerda2004finite} with an alternating checkerboard update scheme for efficient parallelization. The numerical PDE solve times\footnote{All computations were performed on an offboard PC with an AMD Ryzen 9 9950x CPU, coupled with a GeForce RTX 4070 GPU.} ranged from 20~ms to 100~ms. The SQP problem in \eqref{eq: NMPC} was solved at 100~Hz using {\tt OSQP}.

\subsection{Scenario 1: Dynamic Collision Avoidance}

In the first scenario, we commanded both robots (Go2 quadruped and G1 humanoid) to maintain a fixed goal state while simultaneously avoiding an incoming dynamic object, i.e. a red dodgeball. This distills the practical scenario in which legged robots must operate safely in unknown dynamic environments. The outcome of a representative experimental trial for each legged robot are depicted in Fig.\;\ref{fig: quadruped experiment photo}, and the corresponding data is displayed in Fig.\;\ref{fig: quadruped experiment data}.

In the figures, it can be observed that, by considering orientation in the predictive safety filter, both robots efficiently avoid the dynamic obstacle via simultaneous translation and rotation. Specifically, the quadruped initiates its avoidance maneuver by aligning its longitudinal axis (major axis) with the predicted trajectory of the incoming obstacle. Similarly, the humanoid aligns its lateral axis (major axis) with the predicted obstacle trajectory. These twisting actions are pivotal in minimizing the amount of translational motion required to maintain safety. This can be confirmed by examining the evaluated value of $h_{\Q\T}$, which remained positive throughout the avoidance maneuver. This implies that the centroid of the robot remained in $\C_{\Q\T}$ and that all points on the robot remained within the original $\C$. These results were consistent and repeatable, as demonstrated in the video. 

\subsection{Scenario 2: Environmental Navigation}

In the second scenario, we tele-operated the G1 humanoid through an unstructured static environment for 100 seconds. The user-controlled goal point, which was prescribed without inherent safety consideration, was passed to our MPC+CBF predictive safety filter to produce safe velocity commands. These commands were in turn tracked by our RL locomotion controller. The results are shown in Fig.\;\ref{fig: humanoid experiment photo} and Fig.\;\ref{fig: humanoid experiment data}.

By examining Fig.\;\ref{fig: humanoid experiment photo}, the benefit of predictive safety filtering becomes readily apparent. The gap between the obstacles is closed while the robot is in its nominal orientation (left image). Without our geometry-aware safe set buffering method, the robot would not attempt to pass through this corridor, resulting in a navigational ``deadlock". However, the geometry-aware safety-critical controller proceeds to open this gap by planning a trajectory that reorients the humanoid (right image), aligning its lateral axis (major axis) with the corridor traversal direction and allowing the robot to effectively reach its goal position. This point can be further elucidated by examining the Poisson safety function manifolds, and corresponding measured heading angles, in Fig.\;\ref{fig: humanoid experiment data}. Furthermore, the value of the Poisson safety function $h_{\Q\T}$ is enforced to be positive, and this positivity is maintained throughout the preponderance of the 100-second experiment. This again confirms that the 2D footprint of the robot remained within $\C$, satisfying safety requirements. 

\section{CONCLUSIONS}

To summarize, we proposed and demonstrated a simple yet effective means of extending Poisson safety functions into higher-dimensional, non-spatial domains, thus enabling the synthesis of a predictive safety filter. We first established a parameterized moving boundary Dirichlet problem for Poisson's equation to generate a time-varying predictive Poisson safety function. Next, we employed the Minkowski difference operation to lift the domain for Poisson's equation into rotational configuration space. Following these domain extensions, we incorporated our Poisson safety function into an MPC+CBF predictive safety filter algorithm for safe navigation and dynamic collision avoidance. We emphasized the real-time utility of our architecture by implementing it on several legged robots across a breadth of navigational tasks.

\bibliography{main}
\bibliographystyle{ieeetr}

\end{document}

%% file: preamble.tex
\IEEEoverridecommandlockouts
\usepackage{amsmath} 
\usepackage{amssymb}  
\usepackage{amsthm}
\usepackage{mathtools}
\usepackage{xcolor}
\usepackage{url}
\usepackage[hidelinks]{hyperref}
\usepackage{algorithm}
\usepackage{algpseudocode}

\newtheorem{definition}{Definition}

\newtheorem{remark}{Remark}

\newcommand{\bs}[1]{\boldsymbol{ #1 }}
\newcommand{\mb}[1]{\mathbf{ #1 }}
\newcommand{\R}{\mathbb{R}}
\newcommand{\Sp}{\mathbb{S}}
\newcommand{\C}{\mathcal{C}}
\newcommand{\Q}{\mathcal{Q}}
\newcommand{\T}{\mathcal{T}}
\newcommand{\Ocq}{\Omega_\Q}

\newcommand{\Oct}{\Omega_\T}
\newcommand{\Ocqt}{\Omega_{\Q\T}}

\newcommand{\re}{\mathbb{R}}

\newcommand{\Cc}{\mathcal{C}}

\newcommand{\Oc}{\Omega}
\newcommand{\Occ}{\overline{\Omega}}

\newcommand{\Sc}{\mathcal{S}}

\newcommand{\bq}{\bs{q}}

\newcommand{\bvv}{\vec{\bs{v}}}

\newcommand{\by}{\bs{y}}

%% file: main.bbl
\begin{thebibliography}{10}

\bibitem{bansal_hj_2017}
S.~Bansal, M.~Chen, S.~Herbert, and C.~J. Tomlin, ``Hamilton-jacobi reachability: A brief overview and recent advances,'' in {\em 2017 IEEE 56th Annual Conference on Decision and Control (CDC)}, pp.~2242--2253, 2017.

\bibitem{singletary2021comparative}
A.~Singletary, K.~Klingebiel, J.~R. Bourne, N.~A. Browning, P.~Tokumaru, and A.~D. Ames, ``Comparative analysis of control barrier functions and artificial potential fields for obstacle avoidance,'' in {\em IEEE/RSJ International Conference on Intelligent Robots and Systems}, 2021.

\bibitem{borrelli_mpcBook_2017}
F.~Borrelli, A.~Bemporad, and M.~Morari, {\em Predictive Control for Linear and Hybrid Systems}.
\newblock Cambridge University Press, 2017.

\bibitem{ames2019control}
A.~D. Ames, S.~Coogan, M.~Egerstedt, G.~Notomista, K.~Sreenath, and P.~Tabuada, ``Control barrier functions: Theory and applications,'' in {\em 2019 18th European Control Conference (ECC)}, pp.~3420--3431, 2019.

\bibitem{grandia2023perceptive}
R.~Grandia, F.~Jenelten, S.~Yang, F.~Farshidian, and M.~Hutter, ``Perceptive locomotion through nonlinear model-predictive control,'' {\em IEEE Transactions on Robotics}, vol.~39, no.~5, pp.~3402--3421, 2023.

\bibitem{rosolia2019learning}
U.~Rosolia and F.~Borrelli, ``Learning how to autonomously race a car: a predictive control approach,'' {\em IEEE Transactions on Control Systems Technology}, vol.~28, no.~6, pp.~2713--2719, 2019.

\bibitem{zhang2020haptic}
D.~Zhang, G.~Yang, and R.~P. Khurshid, ``Haptic teleoperation of uavs through control barrier functions,'' {\em IEEE Transactions on Haptics}, vol.~13, no.~1, pp.~109--115, 2020.

\bibitem{breeden2022space}
J.~Breeden and D.~Panagou, ``Guaranteed safe spacecraft docking with control barrier functions,'' {\em IEEE Control Systems Letters}, vol.~6, pp.~2000--2005, 2022.

\bibitem{molnar2025collision}
T.~G. Molnar, S.~K. Kannan, J.~Cunningham, K.~Dunlap, K.~L. Hobbs, and A.~D. Ames, ``Collision avoidance and geofencing for fixed-wing aircraft with control barrier functions,'' {\em IEEE Transactions on Control Systems Technology}, pp.~1--16, 2025.

\bibitem{zeng_dtcbfMpc_2021}
J.~Zeng, B.~Zhang, and K.~Sreenath, ``Safety-critical model predictive control with discrete-time control barrier function,'' in {\em 2021 American Control Conference (ACC)}, pp.~3882--3889, 2021.

\bibitem{grandia2021multi}
R.~Grandia, A.~J. Taylor, A.~D. Ames, and M.~Hutter, ``Multi-layered safety for legged robots via control barrier functions and model predictive control,'' in {\em 2021 IEEE International Conference on Robotics and Automation (ICRA)}, pp.~8352--8358, IEEE, 2021.

\bibitem{roque2022corridor}
P.~Roque, W.~S. Cortez, L.~Lindemann, and D.~V. Dimarogonas, ``Corridor mpc: Towards optimal and safe trajectory tracking,'' in {\em 2022 American Control Conference (ACC)}, pp.~2025--2032, 2022.

\bibitem{cosner2025dynamic}
R.~K. Cosner, {\em Dynamic Safety Under Uncertainty: A Control Barrier Function Approach}.
\newblock PhD thesis, California Institute of Technology, 2025.

\bibitem{glotfelter2020nonsmooth}
P.~Glotfelter, J.~Cortes, and M.~Egerstedt, ``A nonsmooth approach to controller synthesis for {Boolean} specifications,'' {\em IEEE Transactions on Automatic Control}, pp.~1--1, 2020.

\bibitem{molnar2023composing}
T.~G. Molnar and A.~D. Ames, ``Composing control barrier functions for complex safety specifications,'' {\em IEEE Control Systems Letters}, 2023.

\bibitem{oleynikova2017voxblox}
H.~Oleynikova, Z.~Taylor, M.~Fehr, R.~Siegwart, and J.~Nieto, ``Voxblox: Incremental 3d euclidean signed distance fields for on-board mav planning,'' in {\em IEEE/RSJ International Conference on Intelligent Robots and Systems (IROS)}, pp.~1366--1373, IEEE, 2017.

\bibitem{long2021learning}
K.~Long, C.~Qian, J.~Cort{\'e}s, and N.~Atanasov, ``Learning barrier functions with memory for robust safe navigation,'' {\em IEEE Robotics and Automation Letters}, vol.~6, no.~3, pp.~4931--4938, 2021.

\bibitem{bahati2025poisson}
G.~Bahati, R.~M. Bena, and A.~D. Ames, ``Dynamic safety in complex environments: Synthesizing safety filters with poisson's equation,'' in {\em Robotics: Science and Systems}, May 2025.

\bibitem{ames2017control}
A.~D. Ames, X.~Xu, J.~W. Grizzle, and P.~Tabuada, ``Control barrier function based quadratic programs for safety critical systems,'' {\em IEEE Transactions on Automatic Control}, vol.~62, no.~8, pp.~3861--3876, 2017.

\bibitem{breeden_sampledDataCbfs_2022}
J.~Breeden, K.~Garg, and D.~Panagou, ``Control barrier functions in sampled-data systems,'' {\em IEEE Control Systems Letters}, vol.~6, pp.~367--372, 2022.

\bibitem{agrawal_dtcbf_2017}
A.~Agrawal and K.~Sreenath, ``Discrete control barrier functions for safety-critical control of discrete systems with application to bipedal robot navigation,'' in {\em Proceedings of Robotics: Science and Systems}, (Cambridge, Massachusetts), July 2017.

\bibitem{nguyen_ecbf_2016}
Q.~Nguyen and K.~Sreenath, ``Exponential control barrier functions for enforcing high relative-degree safety-critical constraints,'' in {\em 2016 American Control Conference (ACC)}, pp.~322--328, 2016.

\bibitem{xiao2022high}
W.~Xiao and C.~Belta, ``High-order control barrier functions,'' {\em IEEE Transactions on Automatic Control}, vol.~67, no.~7, pp.~3655--3662, 2022.

\bibitem{taylor2022backstepping}
A.~J. Taylor, P.~Ong, T.~G. Molnar, and A.~D. Ames, ``Safe backstepping with control barrier functions,'' in {\em 2022 IEEE 61st Conference on Decision and Control (CDC)}, pp.~5775--5782, 2022.

\bibitem{isidori1985nonlinear}
A.~Isidori, {\em Nonlinear control systems: an introduction}.
\newblock Springer, 1985.

\bibitem{mestres2025control}
P.~Mestres, Y.~Chen, E.~Dall'anese, and J.~Cort{\'e}s, ``Control barrier function-based safety filters: Characterization of undesired equilibria, unbounded trajectories, and limit cycles,'' {\em arXiv preprint arXiv:2501.09289}, 2025.

\bibitem{osher2004level}
S.~Osher, R.~Fedkiw, and K.~Piechor, ``Level set methods and dynamic implicit surfaces,'' {\em Appl. Mech. Rev.}, vol.~57, no.~3, pp.~B15--B15, 2004.

\bibitem{evans2022partial}
L.~C. Evans, {\em Partial differential equations}.
\newblock American Mathematical Society, 1998.

\bibitem{hieber2005lagrangian}
S.~E. Hieber and P.~Koumoutsakos, ``A lagrangian particle level set method,'' {\em Journal of Computational Physics}, vol.~210, no.~1, pp.~342--367, 2005.

\bibitem{bonaccorsi2001variational}
S.~Bonaccorsi and G.~Guatteri, ``A variational approach to evolution problems with variable domains,'' {\em Journal of Differential Equations}, vol.~175, no.~1, pp.~51--70, 2001.

\bibitem{bogelein2021evolutionary}
V.~B{\"o}gelein, F.~Duzaar, and C.~Scheven, ``Evolutionary problems in non-cylindrical domains,'' in {\em Harnack Inequalities and Nonlinear Operators: Proceedings of the INdAM conference to celebrate the 70th birthday of Emmanuele DiBenedetto}, pp.~43--60, Springer, 2021.

\bibitem{tamas2022model}
T.~G. Molnar, R.~K. Cosner, A.~W. Singletary, W.~Ubellacker, and A.~D. Ames, ``Model-free safety-critical control for robotic systems,'' {\em Robotics \& Automation Letters}, vol.~7, no.~2, pp.~944--951, 2022.

\bibitem{cohen2024rom}
M.~H. Cohen, T.~G. Molnar, and A.~D. Ames, ``Safety-critical control for autonomous systems: Control barrier functions via reduced order models,'' {\em Annual Reviews in Control}, vol.~57, p.~100947, 2024.

\bibitem{mittal2023orbit}
M.~Mittal, C.~Yu, Q.~Yu, J.~Liu, N.~Rudin, D.~Hoeller, J.~L. Yuan, R.~Singh, Y.~Guo, H.~Mazhar, A.~Mandlekar, B.~Babich, G.~State, M.~Hutter, and A.~Garg, ``Orbit: A unified simulation framework for interactive robot learning environments,'' {\em IEEE Robotics and Automation Letters}, vol.~8, no.~6, pp.~3740--3747, 2023.

\bibitem{gu2024advancing}
X.~Gu, Y.-J. Wang, X.~Zhu, C.~Shi, Y.~Guo, Y.~Liu, and J.~Chen, ``Advancing humanoid locomotion: Mastering challenging terrains with denoising world model learning,'' in {\em Robotics: Science and Systems}, 2024.

\bibitem{xiong2024efficient}
Y.~Xiong, C.~Zhou, X.~Xiang, L.~Wu, C.~Zhu, Z.~Liu, S.~Suri, B.~Varadarajan, R.~Akula, F.~Iandola, {\em et~al.}, ``Efficient track anything,'' {\em arXiv preprint arXiv:2411.18933}, 2024.

\bibitem{farneback2003two}
G.~Farneb{\"a}ck, ``Two-frame motion estimation based on polynomial expansion,'' in {\em Image Analysis}, pp.~363--370, Springer Berlin Heidelberg, 2003.

\bibitem{strikwerda2004finite}
J.~C. Strikwerda, {\em Finite difference schemes and partial differential equations}.
\newblock SIAM, 2004.

\end{thebibliography}
